\documentclass[10pt,twocolumn,letterpaper]{article}

\usepackage{cvpr}              % To produce the CAMERA-READY version
\usepackage[accsupp]{axessibility}  % Improves PDF readability for those with disabilities.
\usepackage{graphicx}
\usepackage{amsmath}
\usepackage{amssymb}
\usepackage{booktabs}
\usepackage{mathtools}
\usepackage[ruled,vlined]{algorithm2e}
\usepackage{algorithmic}
\usepackage{dsfont}
\usepackage{multirow}
\usepackage{multicol}

\makeatletter
\def\hlinew#1{%
  \noalign{\ifnum0=`}\fi\hrule \@height #1 \futurelet
   \reserved@a\@xhline}
\makeatother\makeatletter
\def\hlinew#1{%
  \noalign{\ifnum0=`}\fi\hrule \@height #1 \futurelet
   \reserved@a\@xhline}
\makeatother

\usepackage[pagebackref,breaklinks,colorlinks]{hyperref}

\usepackage[capitalize]{cleveref}
\crefname{section}{Sec.}{Secs.}
\Crefname{section}{Section}{Sections}
\Crefname{table}{Table}{Tables}
\crefname{table}{Tab.}{Tabs.}

\def\cvprPaperID{9638} % *** Enter the CVPR Paper ID here
\def\confName{CVPR}
\def\confYear{2022}

\begin{document}

\title{Evaluation-oriented Knowledge Distillation for Deep Face Recognition}

\author{Yuge Huang$^{\ast}$ ~ ~ Jiaxiang Wu\thanks{~equal contribution. $\ddag$ corresponding author.} ~ ~ Xingkun Xu ~ ~ Shouhong Ding$^{\ddag}$\\
Youtu Lab, Tencent  ~ ~ ~ \\
{\tt\small \{yugehuang, willjxwu, xingkunxu, ericshding\}@tencent.com} \\
{\small \url{https://github.com/Tencent/TFace/tree/master/recognition/tasks/ekd}}
}

\maketitle

%%%%%%%%% ABSTRACT
\begin{abstract}
Knowledge distillation (KD) is a widely-used technique that utilizes large networks to improve the performance of compact models.
Previous KD approaches usually aim to guide the student to mimic the teacher's behavior completely in the representation space.
However, such one-to-one corresponding constraints may lead to inflexible knowledge transfer from the teacher to the student, especially those with low model capacities.
Inspired by the ultimate goal of KD methods, we propose a novel Evaluation-oriented KD method (EKD) for deep face recognition to directly reduce the performance gap between the teacher and student models during training.
Specifically, we adopt the commonly used evaluation metrics in face recognition, \textit{i.e.}, False Positive Rate (FPR) and True Positive Rate (TPR) as the performance indicator.
According to the evaluation protocol, the critical pair relations that cause the TPR and FPR difference between the teacher and student models are selected.
Then, the critical relations in the student are constrained to approximate the corresponding ones in the teacher by a novel rank-based loss function, giving more flexibility to the student with low capacity.
Extensive experimental results on popular benchmarks demonstrate the superiority of our EKD over state-of-the-art competitors.
\end{abstract}

%%%%%%%%% BODY TEXT
\section{Introduction}
\label{sec:intro}
\begin{figure}[t!]
  \centering
  \includegraphics[trim={0 0 0 0mm},clip,width=1\linewidth]{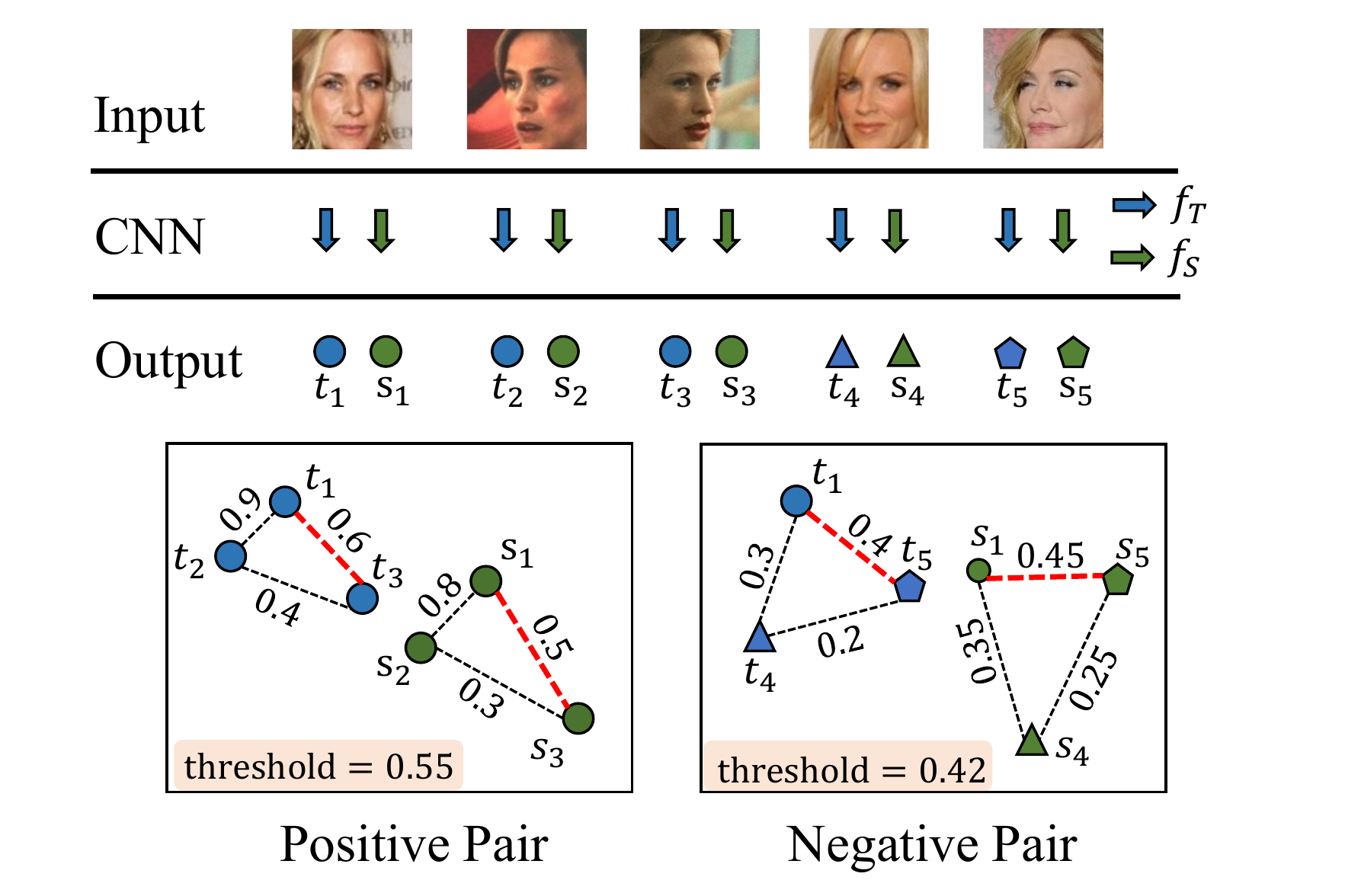}
  \caption{\small Illustration of critical relations of samples.
  Different colors indicate different models (Teacher $T$ in blue and Student $S$ in green). Different shapes indicate samples of different subjects. The numbers denote the cosine similarities of samples.
  The relation of the $1$st and the $3$rd samples is the only one whose similarities fall on the different side of the threshold in teacher and student models (\textit{i.e.}, $0.6 > 0.55$ in teacher while $0.5 < 0.55$ in student), and thus leads to the TPR difference.
  Therefore, in order to pursue the same TPR of the teacher, the student which has limited model capability should pay more attention on the relation (in red) of the $1$st and the $3$rd samples which is the \textbf{critical relation}. Similarly, for the negative pairs, the relation of $1$st and $5$th samples leads to the FPR difference and should be paid more attention.
  } 
  \label{fig:figure1}
\end{figure}
%------------------------------------------------------------------------
With a large number of recognition systems deployed on mobile and edge devices, compact yet discriminative models are in increasingly high demand.
Although some optimized neural network architectures for mobile devices~\cite{chen2018mobilefacenets, sandler2018mobilenetv2} are proposed in the recent years, there still exists an enormous performance gap between these compact networks and the resource-intensive networks which have millions of parameters.
In order to narrow the gap, Knowledge Distillation (KD), which is a widely-used technique that utilizes the knowledge of a large network to improve the performance of the compact models, is proposed.

The seminal works~\cite{2006model_compression, Hinton2014kd} introduced the original idea of KD, which targets on reducing the Kullback–Leibler (KL) divergence between each instance's probabilities at the output layers of the teacher and the student networks.
In the past decade, work~\cite{zagoruyko2016paying, romero2014fitnets, huang2017like} has continued optimizing KD methods by extending such instance-wise constraints to the activation of the hidden layers.
For example, attention transfer~\cite{zagoruyko2016paying} aims to elicit similar response patterns in feature maps.
FitNets~\cite{romero2014fitnets} directly constrains intermediate representations by using regressions.
However, such instance-based methods essentially require the teacher and student to share the same representation space, which is unrealistic for student networks with low model capacities.
As a result, these instance-wised methods bring limited improvement on the performance of student models.
Recently, relation-based KD methods~\cite{park2019relational, tung2019similarity, peng2019correlation} are proposed. Different from the traditional instance-based methods, relation-based ones utilize the correlations between instances as knowledge.
The students in these methods are not required to mimic the teacher's representation space, but rather to preserve the relations of samples in their own representation space.
Thus, they can achieve relatively better performance comparing to the instance-based methods.
However, the model performance trained with these methods are still far from perfect as they still have too strict constraint on knowledge transfer.
In particular, they require the student to mimic all relations between samples in a mini-batch, which seriously limits the flexibility and efficiency of the knowledge transfer from the teacher to the student.

Unlike all the KD methods mentioned earlier, we propose a novel Evaluation-oriented Knowledge Distillation (EKD) method for deep face recognition, which draws inspiration from the ultimate goal of KD, that is, to reduce the performance gap between the teacher and student models. Specifically, we adopt the commonly used evaluation metrics in face recognition, \textit{i.e.}, False Positive Rate (FPR), and True Positive Rate (TPR) as the performance indicator of a face recognition model. 
By performing these two evaluation metrics during the student model training, we can directly obtain the critical pair relations which cause the TPR and FPR difference between the teacher and student models.
Naturally, these critical pairs should be mainly focused on during knowledge transfer.
Thus, we adopt a novel rank-based loss function to constrain the critical relations in the student to approximate the teacher's corresponding ones.
Fig.~\ref{fig:figure1} gives a motivational example and illustrates how critical relations cause the difference of TPR and FPR between the teacher and student models.
Generally, the thresholds of a face recognition model are determined by target FPRs from the similarities of whole negative pairs and are usually different for different models, even if corresponding to the same FPR.
For clarity, we directly give $0.55$ and $0.42$, which roughly correspond to FPR=$1e$-$5$ and FPR=$1e$-$4$, as the thresholds of the student and teacher model.

Although both the proposed {EKD} and the relation-based KD methods optimize the relations between samples, they differ in two aspects.
First, the previous relation-based KD methods require the student to mimic all the relations of the teacher to indirectly reduce the performance gap between the teacher and student models, while our EKD introduces the commonly used evaluation protocol, \textit{i.e.}, TPR and FPR, into the training process and optimizes the critical relations that cause the TPR and FPR difference in the student model to reduce these two metrics gap.
Second, the previous relation-based KD methods usually constrain the absolute similarity of the corresponding pair between the teacher and student models, while our EKD relaxes the constraint by a novel rank-based loss function, which only requires the similarities of the corresponding pairs on the same side of the thresholds in the teacher and student models.

The contributions of this paper are summarized as follows:
\begin{itemize}
\item We propose a novel Evaluation-oriented KD method for deep face recognition. To our best knowledge, EKD is the first KD method to directly reduce the evaluation metric difference between the teacher and student model during training. 
\item We propose a novel rank-based loss function to optimize the student model's critical relations that cause the TPR and FPR difference between the teacher and student models. By only constraining the similarities of the corresponding pairs are on the same side of the thresholds in the teacher and student models, it gives more flexibility to the student, thereby alleviating the student's low capacity problem.
\item We conduct extensive experiments on popular facial benchmarks, which demonstrate the superiority of the proposed EKD over the SOTA competitors.
\end{itemize}

%------------------------------------------------------------------------
\begin{figure*}[t!]
  \centering
  \includegraphics[trim={0 0 0 0mm},clip,width=0.88\linewidth]{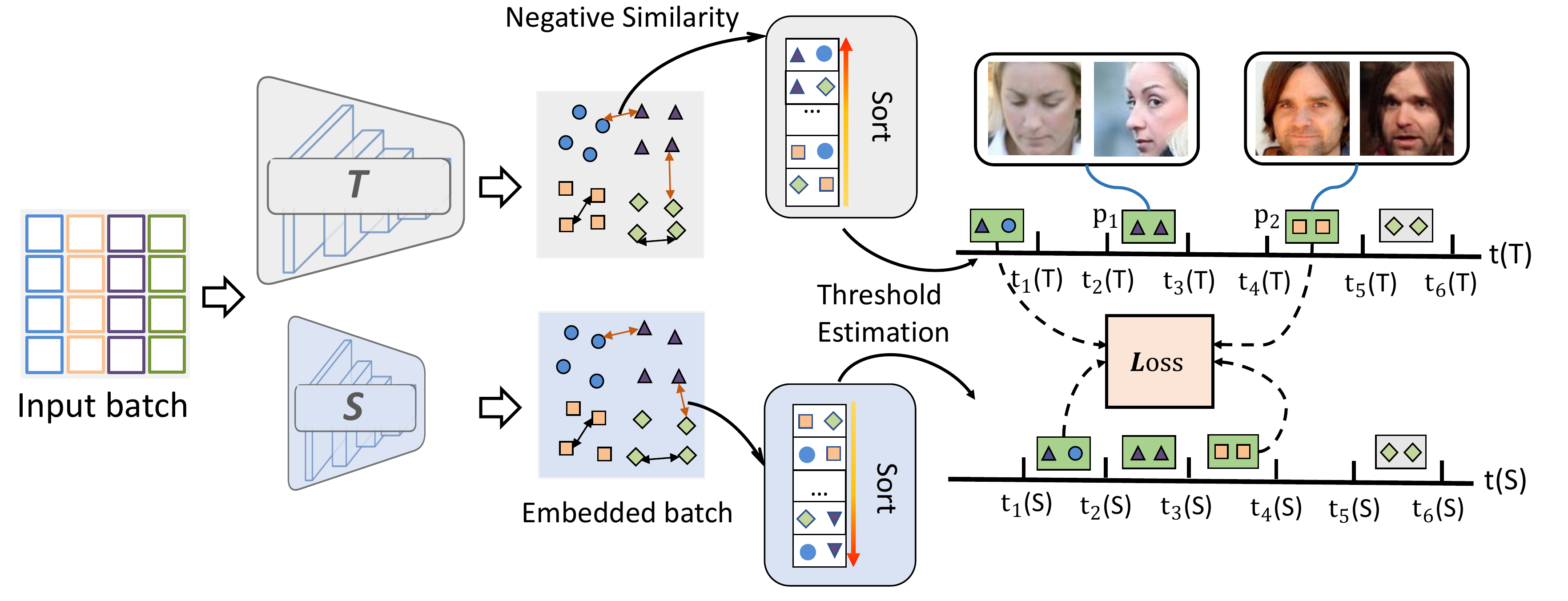}
  \caption{\small Illustration of EKD.
  $T$ and $S$ denote the teacher and student network, $p_1$ and $p_2$ denote the two positive pair relations, respectively.
  The critical pair-wise relations that cause the TPR and FPR difference between the teacher and student model are selected and constrained by the loss function.
  } 
  \label{fig:framework}
\end{figure*}

\section{Related Work}
\label{gen_inst}
\paragraph{Loss Function on Face Recognition.}
Designing a suitable loss function plays a vital role in deep face recognition.
The commonly used loss function can be categorized into two types: metric loss and classification loss. 
Metric losses such as the contrastive~\cite{sun2014deepid2} and the triplet~\cite{schroff2015facenet, Parkhi2015vggface} loss are designed to increase the margin in the Euclidean distance space.
% They usually suffer from high computational cost, thus require carefully-designed sample mining strategies to obtain good performance.
Current SOTA deep face recognition methods mostly adopt softmax-based classification loss~\cite{taigman2014deepface,liu2017sphereface, deng2018arcface, huang2020curricularface}. 
Though such margin-based loss functions equipped with large neural networks are verified to obtain satisfactory performance\cite{deng2018arcface}, they do not always perform well with a mobile neural network~\cite{deng2019lightweight}. The performance gap between the large and compact model motivates us to explore the knowledge distillation method.
%------------------------------------------------------------------------
\paragraph{Knowledge Distillation.} Knowledge distillation has been actively investigated and widely used in many computer vision tasks.
The basic idea proposed by Hinton et al.~\cite{Hinton2014kd} minimizes the KL divergence of soften class probabilities between the teacher and student.
Later, several variants of distillation strategies are proposed to make better use of the teacher network's information.
They mainly fall into two categories, ~\textit{i.e.}, instance-based methods, and relation-based methods. 
Instance-based methods transfer individual outputs from a teacher model to a student model point-wise.
For example, FitNets~\cite{romero2014fitnets} use the intermediate representations of a teacher network to guide the feature activation of a student network.
KD methods especially proposed for face recognition are also mainly in this category.
ShrinkTeaNet~\cite{duong2019shrinkteanet} minimizes the angle of each face sample between teacher and student embedding vectors.
TripletDistillation~\cite{feng2020triplet} improves the triplet loss with dynamic margins by utilizing the similarity structures among different identities in the teacher network.
MarginDistillation~\cite{svitov2020margindistillation} uses class centers from the teacher network for the student network.
Unlike the instance-based methods, relation-based methods~\cite{park2019relational, tung2019similarity, peng2019correlation, chen2017darkrank} transfer relations of the samples in a batch.
RKD~\cite{park2019relational} utilizes two concrete relations, \emph{i.e.}, pairwise and ternary relations of examples. 
SP~\cite{tung2019similarity} and CCKD~\cite{park2019relational} adopt the pairwise similarities of the outputs.
Darkrank~\cite{chen2017darkrank} transfers similarity ranks between data examples.
Although the model performance trained with the two types of KD methods is better than direct training, it is still far from perfect as these methods have too strict constraints on knowledge transfer. 
In particular, instance-based methods require the teacher and student to share the same representation space, while relation-based methods require the student to mimic all relations between samples in a mini-batch.

Our method is related to relation-based methods, but there are several key differences.
Compared with RKD~\cite{park2019relational} and SP~\cite{tung2019similarity}, our method improves in two aspects:
$1$) EKD focuses on the critical relations that cause the TPR and FPR difference between the teacher and student models, while RKD and SP treat all the possible relations equally.
$2$) EKD constrains the critical relations by a novel rank-based loss function to give more flexibility to the student with low capacity, while RKD and SP directly constrain the corresponding similarities by L$2$ loss.
Our EKD and DarkRank~\cite{chen2017darkrank} differ in two aspects:
$1$) EKD adopts the rank between a certain similarity and the thresholds estimated from the total negative pairs in a mini-batch, while DarkRank uses the rank based on the similarity score between the candidate samples and a query sample.
$2$) EKD calculates the rank with an indicator function, which can be simply approximated by a sigmoid function, while DarkRank uses the way introduced by classical list-wise learning to rank methods~\cite{cao2007learning}.
Thus, our method is far simpler to implement.
Besides, the critical relation selection of our method is different from the common hard sample mining strategies in previous methods~\cite{feng2020triplet, lin2017focal}. As illustrated in Fig.~\ref{fig:framework}, the positive pair $p_1$ is more likely to be mined in previous hard sample mining methods. On the contrary, the positive pair $p_2$ is mined in our approach since it leads to the TPR difference between the teacher and student models.
%------------------------------------------------------------------------
\section{The Proposed Method}
Fig.~\ref{fig:framework} illustrates the framework of the proposed EKD. 
Given a teacher model $\mathbf{T}$ and a student model $\mathbf{S}$, we let $f_\mathcal{T}$ and $f_\mathcal{S}$ be functions of the teacher and the student, respectively. 
We follow batch construction from RKD~\cite{park2019relational} and sample $q$ positive images per category in a mini-batch. 
Thus, the features extracted by $T$ and $S$ can be used to construct the positive and negative pairs, respectively.
Then, according to the commonly used evaluation protocol TPR and FPR in face recognition, the critical pair-wise relations that cause the two metrics difference between the teacher and student models are chosen (see Sec.~\ref{relation_selection}). Finally, we constrain the critical relations by a novel rank-based loss function (see Sec.~\ref{ekd}), giving more flexibility to the student and alleviating the student’s low capacity problem.

%------------------------------------------------------------------------
\subsection{Critical Relation Selection}
\label{relation_selection}
%------------------------------------------------------------------------
\paragraph{Positive Pairs and Negative Pairs.}
\label{pair_construction}
First, we introduce the details of constructing the positive and negative pairs in one mini-batch during training. 
A balanced mini-batch consists of $p$ classes, each class with $q$ images.
Therefore, there are $B=p*q$ samples in each mini-batch. 
The number of total pairs are $B*(B-1)/2$, where $p*q*(q-1)/2$ are number of positive pairs and $p*q*(p-1)*q/2$ are negative pairs. 
Following prior art~\cite{deng2018arcface, huang2020curricularface} in face recognition, we adopt the cosine similarity to denote the pair-wise relation:
\begin{equation}
\label{eq:similarity}
\small
    s_{i,j} = \left<f(x_{i}), f(x_{j})\right>, i\ne j
\end{equation}
where $f(x_i)$ denotes the representation of a sample.
%------------------------------------------------------------------------
\paragraph{FPR and TPR Calculation.}
Our method's motivation is directly taking reducing the performance gap between the teacher and the student model as the training constraint.
Thus, the critical problem is to select a suitable evaluation metric as the performance indicator of the model.
In face recognition, TPR and FPR are the most commonly used evaluation metrics.
Thus, we adopt these two evaluation metrics as the model's performance indicator in this work.
We first briefly describe the evaluation protocol of the two metrics.
Given a vector of $M$ similarities $v$ from all the negative pairs, the FPR is computed as the proportion above $t$.
\begin{equation}
\label{eq:fpr}
\small
    FPR(t) = \frac{1}{M}\sum_{i=1}^M \mathds{1}(v_i > t)
\end{equation}
where $t$ is a chosen threshold, $\mathds{1}(x) $ is the discrete Indicator function and $v_i$ denotes the similarity of $i$ relation.
Similarly, given a vector of $N$ genuine scores $u$ from all the positive pairs, the TPR is computed as the proportion above a threshold $t$ as follows.
\begin{equation}
\label{eq:tpr}
\small
    TPR(t) = \frac{1}{N}\sum_{i=1}^N \mathds{1}(u_i > t)
\end{equation}
In practice, the typical way to assess two face recognition models is to fix their FPRs and compare their TPRs. Specifically, the thresholds corresponding to each FPR are determined by the quantiles of all the negative pair similarities, and the TPRs can be calculated from the positive pair similarities based on the obtained thresholds.
The higher the TPRs, the better the model.
The concerned FPR range depends on the deployment scenario of the face recognition system.
For example, the FPR is usually set to be $1e$-$5$ or $1e$-$6$ in a face access control system to balance security and user experience.
In the popular public face benchmarks, the FPR often ranges from $1e$-$1$ to $1e$-$6$ ~\cite{maze2018iarpa, whitelam2017iarpa, kemelmacher2016megaface}.
Thus, we choose [$1e$-$1$, $1e$-$6$] as the target FPR range.
Correspondingly, a vector of $6$ thresholds corresponding to the FPR range evenly spaced on a logarithmic scale can be obtained.
Since the number of negative pairs from one training mini-batch is not large enough, the threshold corresponding to a small FPR value, \textit{e.g.}, $1e$-$6$ may has a large variance. 
We follow~\cite{li2019gradient} to utilize Exponential Moving Average (EMA) to address this issue.
Let $e_k^{n}$ be the estimated $k$-th threshold of the $n$-th batch for the specific FPR and therefore we have:
\begin{equation}
\small{}
\label{eq:threshold_estimation}
t_k = \alpha t_k + (1 - \alpha)e_k^{n},
\end{equation}
where $t_k$ is the $k$-th threshold and initialized with $0$; $\alpha$ is the momentum parameter and set to $0.99$.

\paragraph{Critical Relation Selection.}
According to the above evaluation process, once the thresholds are chosen according to the target FPR ranges, the positive pair relations that cause the TPR difference between the teacher and student model can be obtained.
Though the FPR has been fixed when estimating the corresponding threshold, the difference of the negative pairs in teacher and student models that cause the false positive cases is also instructive during the knowledge transfer.
Thus, the critical relations that cause the difference between the teacher and student models can be defined as follows:
\begin{equation}
\label{eq:condition}
\small
    \mathds{1}(s_{i,j}(T) > t_k(T)) \ne \mathds{1}(s_{i,j}(S)  > t_k(S))
\end{equation}
where $s_{i,j}(T)$ and $s_{i,j}(S)$ denote the similarities between the $i$ and $j$ samples, and $t_k(T)$ and $t_k(S)$ are $k$-th thresholds in the teacher and student models, respectively. The relation between the $i$ and $j$ samples can be positive and negative pairs.
%------------------------------------------------------------------------
\begin{algorithm}[t]
\small
\SetAlgoLined
\KwIn{The balanced input mini-batch $X$, the pre-trained teacher network $T$,
the student network with random initialized parameters $S$,
the FPR range [$FPR_L, FPR_U$], the number of thresholds $k$,
learning rate $\lambda$.}
teacher thresholds $t(T) = [t_1(T), t_2(T), \cdots, t_k(T)] \leftarrow [0, 0, \cdots, 0]$\;
student thresholds $t(S) = [t_1(S), t_2(S), \cdots, t_k(S)] \leftarrow [0, 0, \cdots, 0]$\;
iteration number $i\leftarrow 0$\;
 \While{not converged}{
  Obtain the features by $T$ and $S$\;
  Construct all the possible positive and negative pairs by Eq.~\ref{eq:similarity}\;
  Sort the negative pair similarities and obtain the thresholds corresponding to predefined FPR range in the current mini-batch\;
  Update thresholds $t(T)$ and $t(S)$ by Eq.~\ref{eq:threshold_estimation}\;
  Compute our EKD loss $\mathcal{L}$ by Eq.~$\ref{eq:total_ekd_approximate}$ for positive and negative pairs, respectively\;
  Compute the total loss by Eq.~\ref{eq:total_loss}\;
  Compute the gradients of $S$\;
  Update the parameters $S$\;
  $i \leftarrow i+1$\;
 }
\KwOut{$S$}
 \caption{Evaluation-oriented KD}
 \label{alg:training}
\end{algorithm}

\subsection{Evaluation-oriented Knowledge Distillation}
\label{ekd}
Let $s_{i,j}(T)$ and $s_{i,j}(S)$ denote the similarities between the $i$ and $j$ samples in the teacher and student, respectively. For brevity, the $i$ and $j$ indexes are omitted.
To constrain the critical relations in the student to approximate the corresponding ones in the teacher model, a common loss can be defined as:
\begin{equation}
\label{eq:ekd_hard}
\small
\mathcal{L}_k= \left \lVert s(T) - t_k(T) - (s(S) - t_k(S)) \right \rVert
\end{equation}
where $t_k(T)$ and $t_k(S)$ are the $k$-th thresholds of the teacher and student model, respectively.
Assuming there are $K$ thresholds and $N$ critical relations, the loss function can be formulated as follows:
\begin{equation}
\label{eq:total_ekd_hard}
\small
\mathcal{L}_{hard} = \frac{1}{N}\sum_{n=1}^{N}\sum_{k=1}^{K} \left \lVert s_{n}(T) - t_k(T) - (s_{n}(S) - t_k(S)) \right \rVert
\end{equation}
This formula can be considered as a general loss form used in previous methods like RKD~\cite{park2019relational} and SP~\cite{tung2019similarity}.
If the thresholds of the teacher and student are set to be equal, the loss can be simplified as the common L$2$ loss. 
\begin{equation}
\label{eq:l2}
\small
\mathcal{L}= \frac{1}{N}\sum_{n=1}^{N} \left \lVert s_{n}(T) - s_{n}(S) \right \rVert
\end{equation}

However, the formulation of Eq.~\ref{eq:total_ekd_hard} may still be inflexible due to the absolute distance constraint of each critical relation between the teacher and student models.
Given a positive or negative similarity and a chosen threshold, the comparative relations influence the TPR or FPR rather than the absolute distance.
That is, if a relation meets the condition that $\mathds{1}(s(T) - t_k(T)) = \mathds{1}(s(S) - t_k(S))$, it will not cause the metric difference between the teacher and student models.
Thus, we can directly adopt this condition to optimize the student model.
Since the Indicator function is a step function whose value is $0$ or $1$ and the thresholds are monotonic, the loss can be formulated as follows.
\begin{equation}
\label{eq:total_ekd_soft}
\small
\mathcal{L} = \frac{1}{N}\sum_{n=1}^{N}\left \lVert (\sum_{k=1}^{K}\mathds{1}(s_{n}(T) - t_k(T)) - \sum_{k=1}^{K}\mathds{1}(s_{n}(S) - t_k(S))) \right \rVert
\end{equation}
The above formulation can be considered as a constraint for the rank between a certain similarity and the thresholds.
However, the Indicator function cannot be optimized with gradient-based methods.
Inspired by~\cite{brown2020smooth}, a sigmoid function $G(\cdot;\tau)$ is used to approximate the Indicator function:
\begin{equation}
\label{eq:sigmoid}
\small
\mathcal{G}(x_{nk},\tau)=\frac{1}{1+e^{\frac{-x_{nk}}{\tau}}}
\end{equation}
where $\tau$ refers to the temperature adjusting the sharpness, and $x_{nk}=s_{n}-t_k$ refers to the distance between the $n$-th similarity and the $k$-th threshold.
Substituting $G(\cdot;\tau)$ into Eq.~\ref{eq:total_ekd_soft}, the loss can be approximated as:
\begin{equation}
\label{eq:total_ekd_approximate}
\small
\mathcal{L}_{ekd}= \frac{1}{N}\sum_{n=1}^{N}\left \lVert(  \sum_{k=1}^{K}\mathcal{G}(x_{nk}(T),\tau) - \sum_{k=1}^{K}\mathcal{G}(x_{nk}(S), \tau)) \right \rVert
\end{equation}
where $x_{nk}(T)=s_n(T)-t_k(T)$ and $x_{nk}(S)=s_n(S)-t(S)$.
In addition, as described in Sec.~\ref{pair_construction}, since the number of negative pairs is much larger than the one of positive pairs, we handle the two relations separately and reduce the number of negative pairs via hard negative mining.
In summary, the entire formulation of our EKD is: 
$\mathcal{L}_{EKD} = \lambda_1 \mathcal{L}_{pos} + \lambda_2 \mathcal{L}_{neg}$, where $\lambda_1$ and $\lambda_2$ are the weight parameters.
Furthermore, to maintain the class discriminability, we incorporate the loss function of Arcface~\cite{deng2018arcface}, and thus the final loss becomes: 
\begin{equation}\small
\label{eq:total_loss}
    \mathcal{L}(\Theta) = \mathcal{L}_{EKD} + \mathcal{L}_{Arcface},
\end{equation}
where $\Theta$ denotes the parameter set.
The entire training process is summarized in Algorithm~\ref{alg:training}.

%------------------------------------------------------------------------
\begin{figure}[t!]
  \centering
  \includegraphics[trim={0 0 0 0mm},clip,width=1\linewidth]{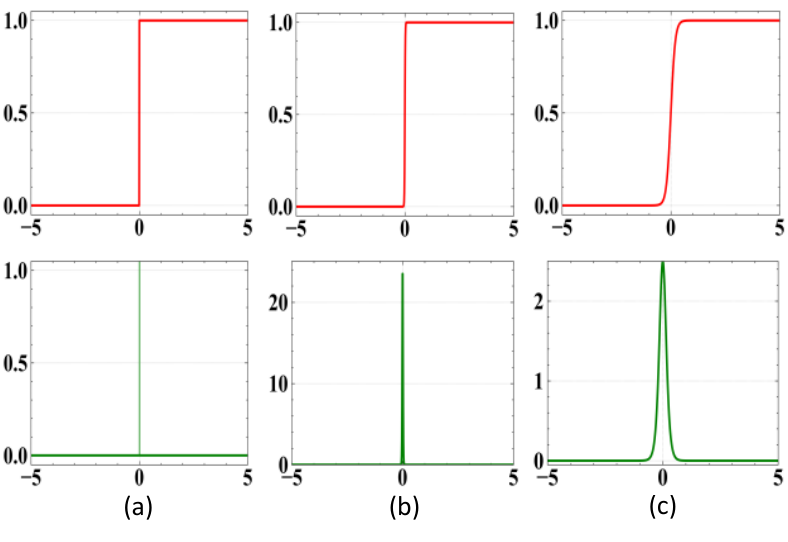}
  \caption{\small (Top) The Indicator function and sigmoid functions with different temperature $\tau$ as different approximations.
  (Bottom) The corresponding derivatives of each function.
  (a) Indicator function (b) sigmoid function with $\tau=0.01$ (c) sigmoid function with $\tau=0.1$.}
  \label{fig:temperature} 
\end{figure}
%------------------------------------------------------------------------
%------------------------------------------------------------------------
\paragraph{Indicator Function Approximation.} The derivative of the Indicator function is defined as Dirac delta function $\delta(x)$, which is either flat everywhere, with zero gradient, or discontinuous, and hence cannot be optimized with gradient based method~\cite{brown2020smooth}.
The derivative of the sigmoid function $\mathcal{G}(x,\tau)$ is as follows:
\begin{equation}\small
    \frac{\partial \mathcal{G}(x,\tau)}{\partial x}=\frac{\mathcal{G}(x,\tau)(1-\mathcal{G}(x,\tau))}{\tau}
\end{equation}
As shown in Fig.~\ref{fig:temperature}, the temperature governs the approximation tightness and the operating region to provide gradients.
%------------------------------------------------------------------------
%-----------------------------------------------------------------------
\section{Experiments}
%-----------------------------------------------------------------------
\subsection{Datasets}
\paragraph{Training Set.} We employ refined MS$1$MV$2$~\cite{deng2018arcface} as our training data for fair comparisons with other methods. MS$1$MV$2$ contains about $5.8$M images of $85$K individuals.
\vspace{-6mm}
\paragraph{Test Set.} We extensively test our method on several popular face benchmarks, including LFW~\cite{lfw}, CFP-FP~\cite{cfp-fp}, CPLFW~\cite{CPLFWTech}, AgeDB~\cite{moschoglou2017agedb}, CALFW~\cite{zheng2017crossage}, IJB-B~\cite{whitelam2017iarpa}, IJB-C~\cite{maze2018iarpa}, and MegaFace~\cite{kemelmacher2016megaface}. 
LFW is the most commonly used face verification test dataset, which contains $13233$ web-collected images from $5749$ different identities. 
The other four datasets are standard benchmarks with two variations, \textit{i.e.}, CFP and CPLFW on pose, and  AgeDB and CALFW on age. 
MegaFace aims at evaluating the face recognition performance at the million scales of distractors. The gallery set of MegaFace includes $1$M images of $690$K subjects, and the probe set includes $100$K photos of $530$ unique subjects from FaceScrub. 
The IJB-B and IJB-C are two challenging public template-based benchmarks for face recognition. 
The IJB-B dataset contains $1,845$ subjects with $21.8$K still images and $55$K frames from $7,011$ videos. 
The IJB-C dataset is a further extension of IJB-B, which contains about $3,500$ identities with a total of $31,334$ images and $117,542$ unconstrained video frames.
%-----------------------------------------------------------------------
\subsection{Experimental Settings}
\paragraph{Data Processing.}
We follow~\cite{deng2018arcface} to crop the $112\times 112$ faces with five landmarks detected by MTCNN~\cite{zhang2016mtcnn}. The RGB images are first normalized by subtracting $127.5$ and divided by $128$, then feeding into the embedding network.
\vspace{-4mm}
\paragraph{Teacher.}
We use Resnet$50$ as the teacher model, which is trained by ArcFace~\cite{deng2018arcface}. For all the experiments in this paper, the teacher model is pre-trained and frozen.
\vspace{-4mm}
\paragraph{Student.} To show our method's generality, we use two neural network structures, \textit{e.g.}, MobileFaceNet~\cite{chen2018mobilefacenets} and Resnet$18$~\cite{deng2018arcface} as the student models, respectively.
\vspace{-4mm}
\paragraph{Training.} We conducted all the experiments on $16$ NVIDIA Tesla V$100$ GPU with Pytorch~\cite{paszke2017automatic} framework.
All student models are trained from scratch using the SGD algorithm for $28$ epochs.
The learning rate starts at $0.1$ and is divided by $10$ at the $10$, $18$, $24$ epochs.
The momentum is $0.9$, and the weight decay is $5e-4$. 
The weights $\lambda_1$ and $\lambda_2$ are set to $0.02$ and $0.01$, respectively.
For ArcFace, we follow the common setting as~\cite{deng2018arcface} to set scale $s=64$ and margin $m=0.5$.
The batch size for ArcFace is set to be $512$.
The balanced batch size is also set to be $512$, and $4$ images are randomly sampled per category.
To increase the number of negative pairs, we merge the two inputs when constructing the negative pairs.
All the training images are horizontally flipped with a probability of $0.5$ as the only data augmentation strategy.
\vspace{-4mm}
\paragraph{Testing.}
We follow the evaluation protocol~\cite{lfw} to report the performance on LFW, CFP-FP, CPLFW, AgeDB and CALFW. 
On Megaface, both face identification and verification performance are reported.
On IJB-B and IJB-C, we follow the $1$:$1$ verification protocol in ArcFace~\cite{deng2018arcface} and take the \textit{average of the image features} as the corresponding template representation without bells and whistles.

\begin{table*}[t!]
\begin{center}
\small
\caption{\small Extensive ablation studies on MS$1$Mv$2$. We report the results of five small test datasets and a large scale test dataset (IJB-C). 
The default student network is MobileFaceNet.
$N$ denotes the number of selected negative pairs.
$K$ denotes the threshold number.
TPR@FPR=$1e$-$4$ and TPR@FPR=$1e$-$5$ on IJB-C are reported.}
\label{tab:ablation_study}
\begin{tabular}{l|c|ccccccc}
\hline
Ablation Type & Methods (\%)      & LFW  & CFP-FP  & CPLFW  & AgeDB   & CALFW  & IJB-C & IJB-C\\ \hline
& ResNet50 (Teacher) & $99.80$ & $97.63$ & $92.50$ & $97.92$ & $96.05$ & $95.16$ & $92.66$ \\\hline
\multirow{4}{*}{Student Structure} & MobileFaceNet     & $99.52$    & $91.66$    & $87.93$   & $95.82$  & $95.12$ & $89.13$   &  $81.65$ \\
& MobileFaceNet + Ours & $\bf{99.60}$    & $\bf{94.33}$    & $\bf{89.35}$   & $\bf{96.48}$  & $\bf{95.37}$ & $\bf{90.48}$ & $\bf{84.00}$ \\
& IR$18$            & $99.67$    & $94.60$    & $89.97$   & $97.33$  & $95.70$   &  $91.96$ & $86.01$ \\
& IR$18$ + Ours     & $\bf{99.68}$    & $\bf{95.31}$    & $\bf{90.82}$   & $\bf{97.48}$  & $\bf{95.85}$   &  $\bf{92.74}$ & $\bf{88.84}$ \\\hline
\multirow{3}{*}{Temperature $\tau$} 
& $\tau = 0.1$        & $99.62$    & $93.33$         & $88.55$        & $96.20$       & $95.20$    &  $89.51$ & $82.04$ \\
& $\tau = 0.01$       & $99.60$    & $\bf{94.33}$    & $\bf{89.35}$   & $\bf{96.48}$  & $\bf{95.37}$   &  $\bf{90.48}$  & $\bf{84.00}$\\
& $\tau = 0.001$      & $\bf{99.65}$   & $93.29$     & $89.07$        & $96.17$                & $95.28$        &  $88.63$ & $79.07$ \\\hline
\multirow{3}{*}{Hard negative mining}
& $N=1000$       & $99.57$       & $93.66$  & $89.28$ & $95.94$ 
    & $95.33$   & $90.29$    & $\bf{84.56}$  \\
& $N=2000$      & $\bf{99.60}$    & $\bf{94.33}$    & $\bf{89.35}$   & $\bf{96.48}$  & $\bf{95.37}$   &  $\bf{90.48}$  & $84.00$\\
& $N=5000$     & $99.58$   & $93.74$     & $88.93$        & $96.35$        
    & $95.30$        & $89.85$  & $82.93$ \\\hline 
\multirow{3}{*}{Random negative selection}
& $N=1000$       & $99.53$   & $94.04$ &  $89.00$   & $96.36$ & $95.10$ & $89.71$ & $83.09$\\
& $N=2000$       & $99.55$   & $94.19$ &  $89.00$   & $96.27$ & $95.33$  & $89.41$  & $82.35$ \\
& $N=5000$       & $99.53$   & $94.17$ &  $89.38$   & $96.15$ & $95.51$ & $89.73$ & $83.02$ \\\hline
\multirow{3}{*}{Threshold Number}
& $K=3$       & $99.53$         & $93.57$         & $88.93$        & $96.05$       & $\bf{95.47}$      & $89.71$      & $83.22$\\
& $K=6$       & $\bf{99.60}$    & $\bf{94.33}$    & $\bf{89.35}$   & $\bf{96.48}$  & $95.37$ &         $\bf{90.48}$ & $\bf{84.00}$ \\\hline
\multirow{2}{*}{Loss function}
& $Eq.~\ref{eq:total_ekd_hard}$        & $99.53$  & $91.99$   & $88.23$  & $96.17$ &  $94.88$ & $89.35$ & $81.68$\\
& $Eq.~\ref{eq:total_ekd_approximate}$  & $\bf{99.60}$    & $\bf{94.33}$    & $\bf{89.35}$   & $\bf{96.48}$  & $\bf{95.37}$ & $\bf{90.48}$ & $\bf{84.00}$
\\\hline
\end{tabular}
\vspace{-6mm}
\end{center}
\end{table*}
%------------------------------------------------------------------------
\subsection{Ablation Study}
%------------------------------------------------------------------------
\paragraph{Effects of Student Network Structure.}
We investigate the generalization capability of our method for different student network structures. Tab.~\ref{tab:ablation_study} (Student Structure) shows the results of two structures, \textit{i.e.}, IR$18$ and MobileFaceNet.
Though the performance improvement on the two network structures is different, our method generally performs better than directly training the student network from scratch.
Our method can bring more improvement for a student with a lower capacity (MobileFaceNet).
%-----------------------------------------------------------------------
\vspace{-4mm}
\paragraph{Effects of the Temperature $\tau$.}
As described in Sec.~\ref{ekd}, the temperature $\tau$ governs the smoothing of the sigmoid function used to approximate the Indicator function. 
Tab.~\ref{tab:ablation_study} (Temperature $\tau$) shows that a value of $0.01$ achieves the best performance, which shares similar conclusion with~\cite{brown2020smooth}.
As shown in Fig.~\ref{fig:temperature}, the value $0.01$ gives a better approximation to the Indicator function than $0.1$ and corresponds to a small operating region to provide gradients.
Though the value of $0.001$ gives a tighter approximation, it cannot provide enough large regions with the gradients.
%-----------------------------------------------------------------------
\vspace{-4mm}
\paragraph{Effects of Hard Negative Mining.}
As describe in Sec.~\ref{ekd}, we adopt the hard negative mining strategy to reduce the number of negative pair similarities.
Firstly, to investigate the influence of the negative pair numbers, we train models with the corresponding strategy ($1000$, $2000$, $5000$ negative pairs with the largest similarity are selected).
The number of positive pairs in a mini-batch is about $800$. Thus we try these values to keep the number of positive pairs and negative pairs comparable.
The comparative results are reported in Tab.~\ref{tab:ablation_study} (Hard negative mining).
We have two observations:
$1$) all the strategies perform better than directly training the student (the row of MobileFaceNet), demonstrating our method's effectiveness.
$2$) The performance of $1000$ and $2000$ is similar, and $5000$ is inferior to the other two.
The reason may be that with the number of negative pairs increasing, the positive pairs' relative weight decreases.
We choose $2000$ as the default value since it achieves the best average performance.
Second, we also investigate the effect of the hard negative mining strategy by replacing it with random negative selection. 
Comparing the results between "Random negative selection" and "Hard negative mining" in Tab.~\ref{tab:ablation_study}, our hard negative mining versions generally perform better than the random selection versions.

%------------------------------------------------------------------------
\vspace{-5mm}
\paragraph{Effects of Thresholds Number.}
Given a concerned FPR range [$FPR_L$, $FPR_U$], the number of thresholds depends on how the FPR is spaced.
In general, a vector of thresholds corresponding to FPR evenly spaced on a logarithmic scale is chosen.
For an FPR range [$1e$-$1$, $1e$-$6$], the typical number of thresholds is $6$. Here, we compare two values, \textit{i.e.}, $3$, $6$.
The $3$ thresholds are set to be as $1e$-$1$, $1e$-$3$, and $1e$-$6$, respectively. 
As shown in Tab.~\ref{tab:ablation_study}, the results of $6$ thresholds are better than $3$, since more thresholds can more finely describe the relations between similarities.

\vspace{-4mm}
\paragraph{Effects of Loss Function.}
To investigate the effect of our relaxed loss function, we train models with Eq.~\ref{eq:total_ekd_hard} and Eq.~\ref{eq:total_ekd_approximate}, respectively. 
Comparing the results in Tab.~\ref{tab:ablation_study} (Loss Function), the model trained with Eq.~\ref{eq:total_ekd_approximate} outperforms the version with Eq.~\ref{eq:total_ekd_hard}, which demonstrates that giving more flexibility to the student is beneficial. 

\begin{figure}[t!]
  \centering
  \includegraphics[trim={0 0 0 0mm},clip,width=0.7\linewidth]{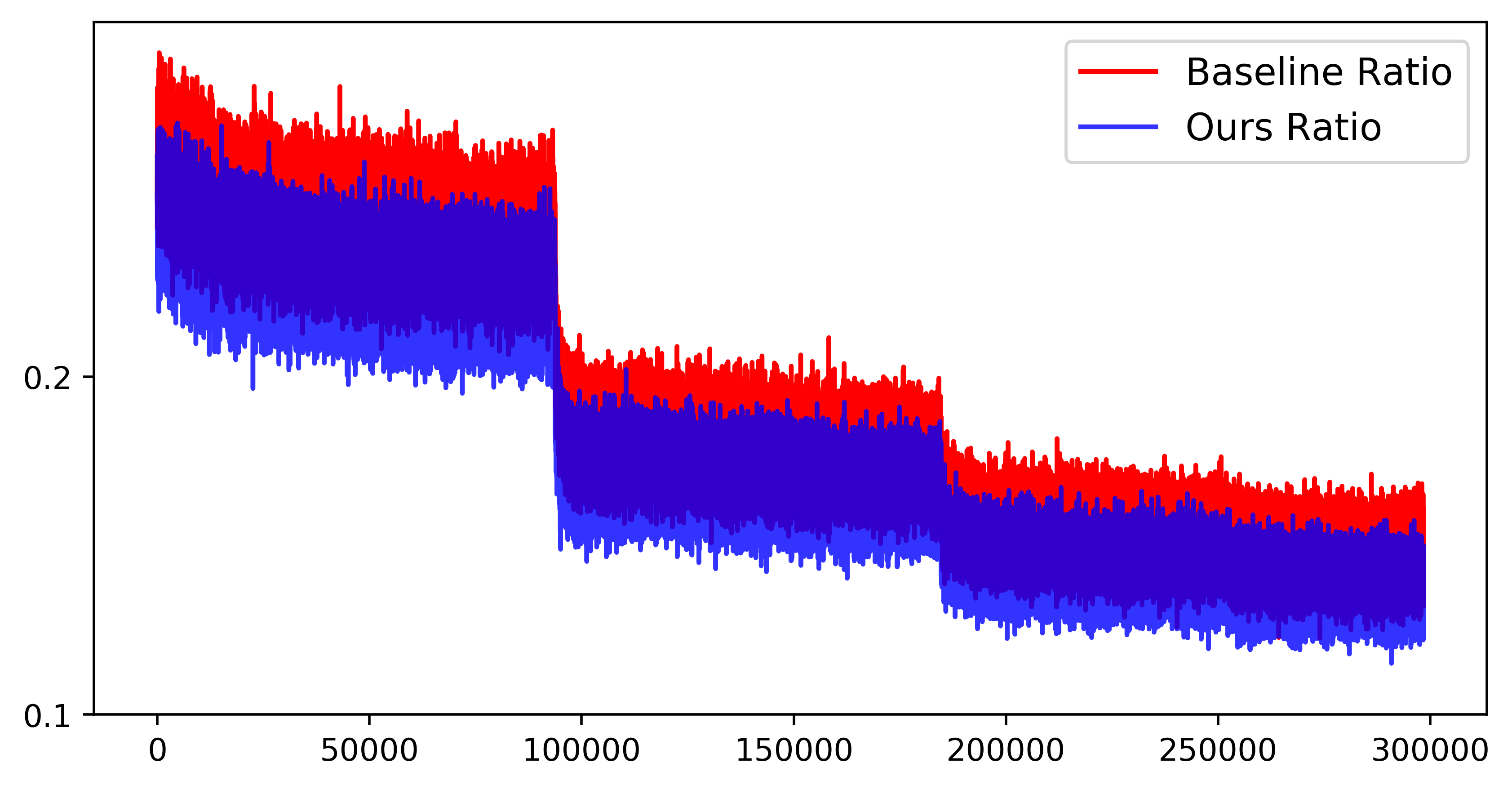}
  \caption{\small Ratio change between the number of critical relations and total relations during training in baseline and ours.}
  \label{fig:critical_ratio}
  \vspace{-4mm}
\end{figure}

\vspace{-2mm}
\paragraph{Ratio between the Number of Critical Relations and Total Relations.}
Fig.~\ref{fig:critical_ratio} shows the ratios between the number of critical relations and total relations, which are calculated during training in the baseline and our method, respectively.
Since the student network is trained from scratch, the ratios at the very early training steps fluctuate wildly, and thus we remove the beginning training steps to make the figure clear.
The number of critical relations trained by our method is smaller than the baseline, demonstrating that our approach does reduce the performance gap between the teacher and student models during training.

%------------------------------------------------------------------------
\subsection{Comparisons with SOTA Methods}
We compare with a wide variety of SOTA KD methods, including the methods proposed for other tasks (
FitNet~\cite{romero2014fitnets}, KD~\cite{Hinton2014kd}, DarkRank~\cite{chen2017darkrank}, SP~\cite{tung2019similarity}, CCKD~\cite{park2019relational} and RKD~\cite{park2019relational}) and specifically designed methods for face recognition (ShrinkTeaNet~\cite{duong2019shrinkteanet}, Triplet Distillation~\cite{feng2020triplet} and MarginDistillation~\cite{svitov2020margindistillation}). 
Since the former six methods do not conduct complete experiments on face recognition, we re-implement them following their original papers.
We cite the results of the latter three methods from~\cite{svitov2020margindistillation}.

\begin{table}[t!]
\begin{center}
\small
\caption{\small Verification comparison with SOTA methods on LFW, two pose benchmarks: CFP-FP and CPLFW, and two age benchmarks: AgeDB and CALFW.}
\label{tab:small_dataset}
\resizebox{1\columnwidth}{!}{
\begin{tabular}{l|ccccc}
\hline
Methods (\%)    & LFW     & CFP-FP  & CPLFW    & AgeDB   & CALFW \\ \hline
ResNet50            & $99.80$ & $97.63$  & $92.50$  & $97.92$ & $96.05$ \\
MobileFaceNet   & $99.52$ & $91.66$  & $87.93$  & $95.82$ & $95.12$ \\\hline
FitNet (arxiv'$14$) & $99.47$ & $91.30$ & $88.30$ & $96.18$ & $95.12$  \\
KD (NIPSW'$14$) & $99.50$ & $91.71$  & $87.85$  & $95.93$ & $95.03$  \\
DarkRank (AAAI'$18$) & $99.55$ & $91.84$  & $87.77$  & $95.60$ & $95.07$  \\
SP (ICCV' $19$) & $99.53$ & $92.33$  & $88.45$  & $96.17$ & $95.07$  \\
CCKD (ICCV' $19$) &$99.47$ & $91.90$ & $88.48$  & $95.83$ & $95.22$ \\
RKD (CVPR'$19$) & $99.58$ & $92.13$  & $87.97$  & $96.18$ & $95.25$  \\
\hline
ShrinkTeaNet (arxiv'$19$) & $99.47$ & $91.97$   & $88.52$ & $96.00$ & $94.98$   \\
TripletDistillation (ICIP'$20$) & $99.55$ & $93.14$ & $88.03$ &$95.53$ & $94.97$ \\
MarginDistillation (arxiv'$20$)  & $\bf{99.61}$ & $92.01$ & $88.03$ &$\bf{96.55}$ & $95.13$  \\\hline
\textbf{EKD (Ours)}  & $99.60$ & $\bf{94.33}$ & $\bf{89.35}$  & $96.48$ & $\bf{95.37}$\\\hline
\end{tabular}
}
\vspace{-5mm}
\end{center}
\end{table}

\begin{figure}[t!]
  \centering
  \includegraphics[trim={0 0 0 0mm},clip,width=0.98\linewidth]{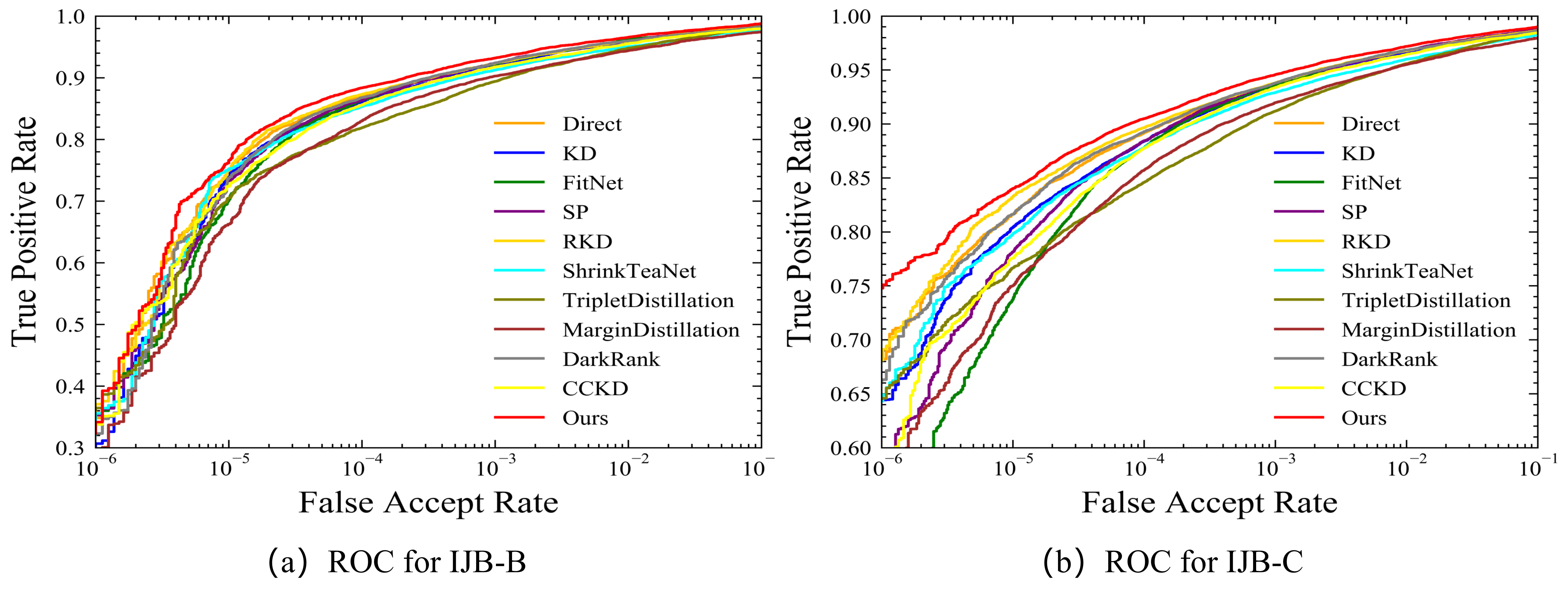}
  \caption{\small ROC curves of 1:1 verification protocol on the IJB-B and IJB-C dataset.}
  \label{fig:comp_ijb}
\end{figure}

\vspace{-4mm}
\paragraph{Results on LFW, CFP-FP, CPLFW, AgeDB and CALFW.}
Tab.~\ref{tab:small_dataset} shows the results compare with the SOTA competitors on five common small benchmarks.
From the Tab.~\ref{tab:small_dataset}, most of the knowledge distillation methods are better than directly training the student network from scratch (\textit{i.e.}, MobileFaceNet), but the performance improvement is limited.
Among all the competitors, relation-based methods seem to show better performance than the instance-based methods, while are inferior to MarginDistillation.
Although we cannot beat the competitors on each test set, we achieve the best average performance on theses test sets.
\vspace{-5mm}
%------------------------------------------------------------------------
%------------------------------------------------------------------------
\paragraph{Results on IJB-B and IJB-C.}
In Tab.~\ref{tab:comp_ijb}, we compare the $1$:$1$ verification TPR@FPR=$1e$-$4$ and TPR@FPR=$1e$-$5$ with the previous SOTA methods on the IJB-B and IJB-C datasets. 
Surprisingly, unlike the small test dataset results, most of the knowledge distillation methods bring a little performance improvement or even worse than the baseline on these two large scale test datasets.
Though RKD shows better generalization ability than others, our method again achieves the best performance.
Fig.~\ref{fig:comp_ijb} shows the full ROC curves of our method and other SOTA competitors, and it is clear that our method performs best.

%------------------------------------------------------------------------
\begin{table}[t!]
\begin{center}
\small
\caption{\small $1$:$1$ verification performance (TPR) on the IJB-B and IJB-C datasets.}
\label{tab:comp_ijb}
\resizebox{1\linewidth}{!}{
\begin{tabular}{l|cccc}
\hlinew{1.2pt}
\multirow{2}{*}{Methods  (\%) }        &  \multicolumn{2}{c}{IJB-C (FPR)} & \multicolumn{2}{c}{IJB-B (FPR)}\\ 
                         \cmidrule(r){2-3} \cmidrule{4-5}
                         &$1e$$-$$4$ & $1e$$-$$5$  
                         &$1e$$-$$4$ & $1e$$-$$5$  \\
                         \hline
ResNet50                 & $95.16$     & $92.66$  &  $93.45$   & $88.65$ \\
MobileFaceNet        & $89.13$     & $81.65$  &  $87.07$   & $74.63$  \\\hline
FitNet (arxiv'$14$)  & $87.76$    &  $73.71$  & $86.35$    & $70.19$ \\
KD (NIPSW'$14$)      & $88.37$    &  $80.39$  & $86.08$    & $74.30$ \\
DarkRank (AAAI'$18$) & $89.28$    &  $81.62$  & $86.76$    & $73.75$ \\
SP (ICCV' $19$)      & $88.43$    &  $78.13$  & $86.34$    & $72.85$ \\
CCKD (ICCV' $19$)    & $87.99$    &  $78.75$  & $85.63$    & $72.38$
\\
RKD (CVPR'$19$)      & $89.65$    &  $83.21$  & $87.27$    & $75.17$ \\\hline
ShrinkTeaNet (arxiv'$19$)         & $87.80$    &  $79.78$      & $85.31$    & $75.23$ \\
TripletDistillation (ICIP'$20$)  & $84.57$    &  $76.65$      & $81.88$    & $70.51$ \\
MarginDistillation (arxiv'$20$)   & $85.71$    &  $75.00$      & $82.97$    & $66.25$ \\\hline
\textbf{EKD (Ours)}              & $\bf{90.48}$ & $\bf{84.00}$ & $\bf{88.35}$ & $\bf{76.60}$ \\\hline
\end{tabular}
}
\vspace{-4mm}
\end{center}
\end{table}
%------------------------------------------------------------------------
%------------------------------------------------------------------------
\vspace{-5mm}
\paragraph{Results on MegaFace.}
Finally, we evaluate the performance on the MegaFace Challenge.
We also report the results following the ArcFace testing protocol, which refines both the probe set and the gallery set.
As shown in Tab.~\ref{tab:comp_megaface}, most of the competitors achieve better performance than baseline, and our method achieves the best verification performance, surpassing all the other strong competitors.
Our method performs slightly inferior to the others on the rank-$1$ metric.
The reason may be that our method adopts the TPR and FPR as the performance indicator during training and overlooks the top$1$ performance.
\vspace{-4mm}
\paragraph{Time Complexity.}
As shown in Tab.~\ref{tab:training_time}, though our method brings some burden on training complexity compared with directly training the small network without the teacher, our method has lower complexity compared with RKD, which is also a relation-based KD method.

%------------------------------------------------------------------------
\begin{table}[t!]
\begin{center}
\small
\caption{\small Verification comparison with SOTA methods on MegaFace Challenge $1$ using FaceScrub as the probe set. “Id” refers to the rank-$1$ face identification accuracy with $1$M distractors, and “Ver” refers to the face verification TPR at $1e$-$6$ FPR. The column “R” refers to data refinement on both probe set and $1$M distractors.}
\label{tab:comp_megaface}
\resizebox{1\columnwidth}{!}{
\begin{tabular}{l|cccc}
\hline
Methods (\%) &  Id (R)  & Ver (R)  & Id & Ver \\ \hline\hline
ResNet50                 & $98.14$     & $98.34$   &  $80.62$  & $96.83$ \\
MobileFaceNet        & $90.91$     & $92.71$   &  $75.52$  & $90.80$  \\\hline
FitNet (arxiv'$14$)  & $91.16$    &  $92.34$   & $75.88$    & $90.64$ \\
KD (NIPSW'$14$)      & $90.40$    &  $92.00$   & $75.81$    & $90.07$ \\
DarkRank (AAAI'$18$) & $90.76$    &  $92.41$   & $75.80$    & $90.66$ \\
SP (ICCV' $19$)      & $91.25$    &  $92.41$   & $75.37$    & $90.62$ \\
CCKD (ICCV' $19$)    & $91.17$    &  $92.76$   & $75.73$    & 
$90.63$\\ 
RKD (CVPR'$19$)      & $91.44$    &  $92.92$   & $75.73$    & $91.21$ \\\hline
ShrinkTeaNet (arxiv'$19$)         & $90.73$    &  $92.32$   & $75.55$    & $90.56$ \\
Triplet Distillation (ICIP'$20$)  & $86.52$    &  $88.75$      & $71.93$    & $91.35$ \\
MarginDistillation (arxiv'$20$)   & $\bf{91.70}$    &  $92.96$      & $\bf{76.34}$    & $91.31$ \\\hline
\textbf{EKD (Ours)}              & $91.02$ & $\bf{93.08}$ & $75.54$ & $\bf{91.42}$ \\\hline
\end{tabular}
}
\vspace{-4mm}
\end{center}
\end{table}

\begin{table}[t!]
\begin{center}
\small
\caption{Training time for each batch under the same experiment setting.}
\vspace{-5mm}
\label{tab:training_time}
\resizebox{0.65\columnwidth}{!}{
\begin{tabular}{l|ccc}
\hline
Methods & Baseline & RKD & Ours \\ \hline
Time (s) & 0.068 & 0.147 & 0.129 \\ \hline
\end{tabular}
}
\vspace{-8mm}
\end{center}
\end{table}
%------------------------------------------------------------------------
%------------------------------------------------------------------------
\section{Conclusions}
In this paper, we propose a novel evaluation-oriented KD method for deep face recognition.
Different from previous KD methods requiring the student to mimic the teacher's behavior completely in the representation space, our EKD optimizes the student to directly reduce the performance gap between the teacher and student models during training.
The most commonly used evaluation metrics in face recognition, \textit{i.e.}, False Positive Rate (FPR), and True Positive Rate (TPR),  are adopted as the performance indicator.
Extensive experiments on popular face recognition benchmarks have demonstrated our method's effectiveness and generalization capability.
In subsequent work, we can try to improve the TOP$1$ performance of our method with more suitable performance evaluation metrics. 

%%%%%%%%% REFERENCES
{\small
\bibliographystyle{ieee_fullname}
\bibliography{egbib}
}

\end{document}